\documentclass{article}
\usepackage{spconf,amsmath,graphicx}
\usepackage{times}
\usepackage{epsfig}
\usepackage{graphicx}
\usepackage{amsmath}
\usepackage{amssymb}
\usepackage{tabularx}
\usepackage{algorithmic}
\usepackage{algorithm}
\usepackage{subcaption}
\usepackage{textcomp}
\usepackage{placeins}
\usepackage{epstopdf}
\usepackage{multirow}
\usepackage{siunitx}
\usepackage{makecell,multirow}
\usepackage{subcaption}
\usepackage[utf8]{inputenc}
\usepackage{hyperref}

\title{TRANSFER LEARNING USING CLASSIFICATION LAYER FEATURES OF CNN}
\name{Tasfia Shermin \quad Manzur Murshed \quad Guojun Lu \quad Shyh Wei Teng\thanks{}}
\address{School of Science, Engineering and Information Technology\\
Federation University Australia, Gippsland Campus, Churchill, VIC 3842, Australia }
\begin{document}
%
\maketitle
\begin{abstract}
Although CNNs have gained the ability to transfer learned knowledge from source task to target task by virtue of large annotated datasets but consume huge processing time to fine-tune without GPU. In this paper, we propose a new computationally efficient transfer learning approach using classification layer features of pre-trained CNNs by appending layer after existing classification layer. We demonstrate that fine-tuning of the appended layer with existing classification layer for new task converges much faster than baseline and in average outperforms baseline classification accuracy. Furthermore, we execute thorough experiments to examine the influence of quantity, similarity, and dissimilarity of training sets in our classification outcomes to demonstrate transferability of classification layer features.
\end{abstract}
\begin{keywords}
Transfer learning, deep networks, computational efficiency, classification
\end{keywords}
\section{Introduction}
\label{sec:intro}
The advancement of influential internal representations in human infancy is reused later in life to solve various problems as stated by the cognitive study of \cite{Alpher02}. In resemblance to humans, deep neural networks built for computer vision problems also learn the data representations (features) which they use later to solve multiple tasks. This phenomenon of transferability of learned data representations is termed as transfer learning \cite{caruana1995learning,bengio2012deep,bengio2011deep}. This technique works well when the learned features are generic, which refers to having features suitable to both base and target datasets. The opportunity to learn generic features for deep networks is paved by the ImageNet \cite{deng2009imagenet} dataset. Deep neural networks incline to learn generic features in the first layer that resemble Gabor filters and colour blobs irrespective of datasets and training objectives \cite{krizhevsky2012imagenet,le2011ica,lee2009convolutional}. A number of works in various computer vision tasks have reported significant results by transferring inner layer features of deep networks \cite{donahue2013deep,zeiler2014visualizing,sermanet2013overfeat}. As the deep network architecture moves toward fully-connected (FC) layers, the specificity increases while the generic nature of features decrease \cite{yosinski2014transferable}, \textit{i.e.}, the intuition is that they are highly specific to pre-trained classes and might not generalize well in transferring knowledge. 
\begin{figure}
   \includegraphics[width=.5\textwidth,height=\textheight,keepaspectratio]{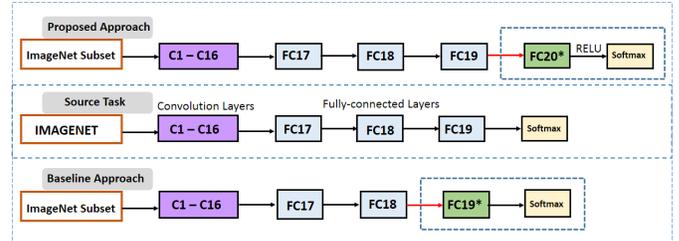}
   \caption{Both proposed and baseline approach of transfer learning with corresponding source task are shown for VGG19 network architecture}
   \label{flow}
\end{figure}
 However, a recent research by \cite{wang2017growing} demonstrated CNN architectures either by adding FC layers in between final FC layer and classification layer or by widening existing FC layers which outperformed classic fine-tuned transfer learning. They have used step wise hyper-parameters to keep pace with training time constraints. Motivated by their findings, we hypothesize to append new FC layer after existing classification layer (\textit{e.g. FC19} for VGG19) as shown in Fig.\ref{flow}. Eventually, we fine-tune new and existing classification layer to investigate that proposed approach consumes less training time because of having only 1000 dimensional feature vectors. In addition, do not adversely affect classification accuracy. Performance of proposed approach is compared with an existing approach which replaces the final FC layer with the number of new classes and fine-tune penultimate layer consisting 4096 or higher feature vectors along with the replaced one. Furthermore, we systematically investigate the following research questions (RQ) to study the impact of training sets in both proposed and baseline approaches. Eventually, demonstrate that classification layer features have similar behaviour as other FC layers for transfer learning.

 {\bf RQ1:} Does similarity of new classes with the pre-trained classes influence performance of classification using transfer learning?
 
 {\bf RQ2:} Does similarity among new classes influence the performance of classification using transfer learning?
 
 {\bf RQ3:} How much the performance of classification using transfer learning is influenced by the number of training and validation images used for new classes?
 
 {\bf RQ4:} How much the performance of classification using transfer learning is influenced when a mixed types of new classes are trained?
 
 {\bf RQ5:} Can proposed approach be used to improve computational efficiency without adversely affecting the performance of classification?

\section{Related Work}
\label{sec:format}
A significant number of papers have experimented and studied transfer learning in CNNs, which includes various factors affecting fine-tuning, pre-training and freezing layers. Apparently, it has become a trend for computer vision community to treat convolutional neural networks \cite{krizhevsky2012imagenet,lecun1998gradient,szegedy2015going,simonyan2014very,he2016deep} trained on ImageNet as extractors of features that can be reused in handling visualization tasks. Discussion on whether to stop pre-training early to avoid overfitting and which layers would be best transferable for transfer learning is studied by \cite{agrawal2014analyzing,yosinski2014transferable}. \cite{sun2017revisiting} have investigated transfer learning based on noisy data. Fine-tune for new tasks without forgetting the old ones is proposed by \cite{li2017learning}. To limit the need for annotated data required for transfer learning, \cite{tamaazousti2017learning} has proposed a method of more universal representations. The nature of transfer learning with mid-level features is studied by \cite{oquab2014learning}. CNN features were used as off-the-shelf features by \cite{sharif2014cnn}. CNN features pre-trained in road scenes were reused for more specific road scene classifications by \cite{holder2016road}.\\
In this paper, we propose a transfer learning approach using pre-trained classification layer's output feature which converges faster during fine-tuning. We have used hyper-parameters and activation function for fine-tuning appended layer based on its fully-connected structure, which fuels computational efficiency and yields competitive results to the baseline. Finally, investigate effect of training sets in our outcomes to give evidence of transferability of classification layer features.

\section{Proposed technique and baseline}
\label{sec:pagestyle}
 We have used ImageNet-1000 pre-trained CNNs \cite{simonyan2014very,he2016deep} for both proposed and baseline approach. For augmenting the training dataset, input images from training sets are first randomly cropped, horizontally flipped (randomly) and then normalized. The pre-trained base networks are designed to take square images as inputs (\textit{i.e.}, $\textit{H\textsubscript{I}} = \textit{W\textsubscript{I}}$). Therefore, to match the input dimension of the network, square patches \textit{S} of (\textit{maximum}) height and width $\min(\textit{H\textsubscript{I}}, \textit{W\textsubscript{I}})$ are randomly cropped from the image. The cropped patches are then resized to $\textit{H\textsubscript{S}} \times \textit{W\textsubscript{S}} \times \textit{D\textsubscript{s}}$ preserving the aspect ratio of the image. For validation and testing, center instead of random crop of the image is taken followed by resizing.\\
 For baseline, the classification layer is replaced with the number of classes in the target experiment \cite{oquab2014learning, sharif2014cnn}. The baseline is represented in Fig.\ref{flow}. The classification and the FC layers with 4096 or more feature vectors are fine-tuned for 25 epochs with a learning rate of $10^{-3}$. Stochastic gradient descent (SGD) \cite{bottou2010large} is used with a momentum of 0.9 and no decaying of weight. During training, cross-entropy loss is used and scheduler step size being set to 7 with a gamma value which equals to $10^{-1}$.\\
  For proposed approach, each of the 1000 neurons of ImageNet pre-trained CNN is connected to every neurons in the newly added layer. Number of neurons in new classification layer is decided according to the number of classes in the target task. For introducing non-linearity in the model, we have activated the neurons of the new layer with Rectified Linear Unit (RELU) \cite{nair2010rectified}. Our empirical study indicates that for fine-tuning, the initialization of learnable weights and biases following uniform distribution yields best results. The values of weight and bias are initialized from
$\mathcal{U}(-\sqrt{k}, \sqrt{k})$, where $k = {1}/{\textit{No. of inputs.}}$ Stochastic gradient descent (SGD) optimization was used with a learning rate of $10^{-2}$ and no momentum. The learning rate was decayed by a factor of $10^{-1}$ after every 7 epochs. For the purpose of calculating loss function, categorical cross entropy loss was used.
\section{Experimental Setup}
\label{sec:typestyle}
Following our questions of interest stated in Section 1, four types of species (\textit{Bird}, \textit{Fruit}, \textit{Flower} and \textit{Pepper}) consisting five different classes each with different degrees of similarity (80\%, 70\%, 60\%, and 50\%) approx. to the pre-trained classes have been selected. The percentage of similarity of new class with respect to pre-trained class of a species is tested based on the output of the pre-trained network. 500 images for each classes of the species were collected according to the ImageNet synsets by web crawling. The target and base datasets had no overlapping classes. For comparing proposed and baseline approaches to classify among different classes of same species, three types of classification (\textit{i.e.}, 3-class, 4-class and 5-class) \textit{A-type classification} are performed. For investigating, classification among different classes of different species \textit{B-type classification}, experiments are designed with ${\textit{k}}/{4}$ classes from each species, where $\textit{k} \in \{12\}$. We have used three combinations of target sets consisting $\textit{fJ}$ images for training, ${\textit{fJ}}/{2}$ images for validation, and $({1-3\textit{f}}/{2}){\textit{J}}$ images for testing from each class, where $\textit{J} = {500}$ and $\textit{f} \in \{10\%, 20\%, 40\%\}$. For example, the first target set is composed of 50 training images, 25 validation images and the rest of the images are left for testing from each classes. The retrieval of pre-trained weights and other experiments are done in PyTorch. Hyper-parameters of all experiments were tuned by 30-fold cross-validation. Two different performance metrics are considered: test accuracy (TA) obtained on the test sets and training time (TT) of the networks. 
\begin{table}[!tb]
\caption{TA (\%) of the proposed approach against the baseline for training each species independently or in a mix with fixed number of classes per species.}
		\label{tab:ind_mix}
	\begin{center}
		\begin{tabular}{|c|c|ccc|}
			\hline
			\multirow{2}{*}{Species} & \multirow{2}{*}{CNN} & \multicolumn{3}{c|}{3 classes per species} \\ \cline{3-5} 
			&  & Baseline & P & Gain \\ \hline \hline
			\multirow{3}{*}{Indep. (avg)} & ResNet18 & 76.2 & 77.3 & 1.1\% \\
			& VGG19 & 75.9 & 76.2 & 1.0\% \\ \cline{2-5} 
			& Average & 76.1 & 77.0 & 1.0\% \\ \hline
			\multirow{3}{*}{Mixed} & ResNet18 & 71.1 & 72.9 & 1.7\% \\
			& VGG19 & 73.8 & 75.1 & 1.3\% \\ \cline{2-5} 
			& Average & 72.4 & 74.0 & 1.5\% \\ \hline
		\end{tabular}	
	\end{center}

\end{table}

\begin{table}[!tb]
	\caption{Transfer learning TT(s) of the proposed approach against the baseline.}
			\label{tab:TT}
	\begin{center}
	\resizebox{\columnwidth}{!}{%
		\begin{tabular}{|c|c|cc|cc|cc|}
			\hline
			\multirow{3}{*}{CNN} & \multirow{3}{*}{Classes} & \multicolumn{6}{c|}{No. of training images} \\ \cline{3-8} 
			&  & \multicolumn{2}{c|}{50} & \multicolumn{2}{c|}{100} & \multicolumn{2}{c|}{200} \\ \cline{3-8} 
			&  & Baseline & P & Baseline & P & Baseline & P \\ \hline \hline
			\multirow{5}{*}{ResNet18} & 3 & 990 & 15 & 1110 & 17 & 1130 & 18 \\
			& 4 & 1110 & 16 & 1230 & 19 & 1230 & 19 \\
			& 5 & 1808 & 18 & 1832 & 22 & 1868 & 23 \\ \cline{2-8} 
			& Average & 1303 & 16 & 1391 & 19 & 1409 & 20 \\ \cline{2-8} 
			& Gain & \multicolumn{2}{c|}{-98.7\%} & \multicolumn{2}{c|}{-98.6\%} & \multicolumn{2}{c|}{-98.6\%} \\ \hline
			\multirow{5}{*}{VGG19} & 3 & 1215 & 18 & 1315 & 20 & 1325 & 23 \\
			& 4 & 1255 & 22 & 1505 & 23 & 1535 & 25 \\
			& 5 & 1935 & 28 & 2115 & 29 & 2175 & 30 \\ \cline{2-8} 
			& Average & 1468 & 23 & 1645 & 24 & 1678 & 26 \\ \cline{2-8} 
			& Gain & \multicolumn{2}{c|}{-98.5\%} & \multicolumn{2}{c|}{-98.5\%} & \multicolumn{2}{c|}{-98.5\%} \\ \hline
		\end{tabular}
		}
	\end{center}

\end{table}

\section{RESULTS AND DISCUSSION}
\label{sec:majhead}
This section discusses detail in the light of our 5 research questions about the findings from experiments by observing the outcomes portrayed in Tables, where \textit{P} denotes Proposed approach. Percentage of gain of TA denotes difference between proposed and baseline, where negative (-) sign indicates less TA of proposed approach compared to baseline. 

{\bf RQ1:} Classification outcomes using classification layer's output features follow the trend of behaviour of classification in computer vision. From Tables \ref{tab:accuracy50}, \ref{tab:accuracy100} and \ref{tab:accuracy200} it can be stated that for all cases of \textit{ A-type classification} with the gradual diminution of similarity TA for proposed and baseline approach decreases. This observation establishes a relation between similarity of new and pre-trained classes which highly influence classification outcomes.

{\bf RQ2:} Moreover, TA decreases with increment of number of classes. For example, if one observes towards right starting from column 4 in Table \ref{tab:accuracy50} TA of both baseline and proposed technique decreases to $90.0\%$ from $91.0\%$ and $91.4\%$ from $92.1\%$ respectively. This phenomenon indicates the marginal improvement gradually decreases with the increase of classes of same species. Therefore, similarity among new classes of same type does not seem to have much impact in increasing the performance of classification.

{\bf RQ3:} With the increase of No. of training samples per class the performance of TA increases. For example, Table \ref{tab:accuracy50} shows that for Birds (\textit{A-type classification} with 3 classes) TA of proposed approach is $92.1\%$ along with the increase in number of training samples for each class TA in Table \ref{tab:accuracy200} it becomes $93.9\%$. Similar $1\%$ (approx) increase is frequently observed in all species with different similarity which indicates more training samples help to learn more and yields better performance. Apparently, comparison among similarity and TA clearly shows approximately $10\%$ of increase in classification after transfer learning using both approaches. Which establishes classification layer's output features are suitable for \textit{A-type classification} tasks. In addition, proposed approach yields very competitive classification TA by using only 1000 dimensional feature vector compared to baseline technique which uses 4096 or higher dimensional features. Proposed approach achieves average gain in the range of $0.7\%$ to $1.5\%$ as observed from Table \ref{tab:accuracy50}, \ref{tab:accuracy100}, and \ref{tab:accuracy200}. Concerning the TA of proposed technique, it is seen from results that on average it performs similar to baseline and for some cases it outperforms baseline by $1.5\%$ approximately.

{\bf RQ4:} To understand behaviour of mixed species in classification, 3 class {\textit A-type classification} is compared with 3 classes per species for {\textit B-type classification}. From Table \ref{tab:ind_mix} it is apparent that proposed approach achieves more average gain for mixed class experiments. Which establishes that proposed approach does better classification than baseline when similarity among classes decreases with the increase of number of classes.
 
{\bf RQ5:} For providing evidence about computational efficiency, avg. TT of both proposed and baseline techniques are enlisted in Table \ref{tab:TT}. All TT are presented in seconds. For TT, negative (-) gains indicate less time needed to train. It is noticed that for all cases, our approach is approximately $98$ times faster than baseline. This fast training is fuelled by a better initialization, suitable learning rate and faster forward propagation (due to having less fully connected neurons). Proposed network does not overfit because of early stopping at the time of convergence.  

\begin{table*}[!tb]
\caption{TA (\%) of the proposed approach against the baseline for 50 training images per class of each species.}
	\label{tab:accuracy50}
	\begin{center}
		\begin{tabular}{|c|c|c|ccc|ccc|ccc|}
			\hline
			\multirow{3}{*}{Species} & \multirow{3}{*}{CNN} & \multirow{3}{*}{Similarity} & \multicolumn{9}{c|}{Classes} \\ \cline{4-12} 
			&  &  & \multicolumn{3}{c|}{3} & \multicolumn{3}{c|}{4} & \multicolumn{3}{c|}{5} \\ \cline{4-12} 
			&  &  & Baseline & P & Gain & Baseline & P & Gain & Baseline & P & Gain \\ \hline \hline
			\multirow{2}{*}{\textit{Bird}} & ResNet18 & 81.8 & 91.0 & 92.1 & 1.1\% & 90.5 & 91.9 & 1.4\% & 90.0 & 91.4 & 1.4\% \\
			& VGG19 & 80.8 & 92.0 & 91.8 & -0.2\% & 90.0 & 90.7 & 0.7\% & 90.5 & 90.9 & 0.4\% \\
			\multirow{2}{*}{\textit{Fruit}} & ResNet18 & 72.5 & 80.5 & 81.8 & 1.3\% & 80.3 & 81.0 & 0.7\% & 80.2 & 81.0 & 0.9\% \\
			& VGG19 & 72.7 & 79.0 & 79.3 & 0.4\% & 78.3 & 79.0 & 0.9\% & 78.5 & 78.8 & 0.4\% \\
			\multirow{2}{*}{\textit{Flower}} & ResNet18 & 64.8 & 72.3 & 73.0 & 0.9\% & 72.1 & 72.8 & 1.0\% & 72.1 & 72.7 & 0.9\% \\
			& VGG19 & 63.7 & 70.5 & 70.1 & -0.4\% & 70.3 & 70.0 & -0.3\% & 70.3 & 70.0 & -0.3\% \\
			\multirow{2}{*}{\textit{Pepper}} & ResNet18 & 50.9 & 61.2 & 64.2 & 4.7\% & 61.1 & 64.1 & 4.7\% & 61.1 & 64.7 & 5.6\% \\
			& VGG19 & 52.2 & 62.2 & 62.6 & 0.7\% & 62.3 & 62.0 & -0.5\% & 62.3 & 62.0 & -0.5\% \\ \hline
			\multicolumn{2}{|c|}{Average} & 67.4 & 76.1 & 76.9 & 1.1\% & 75.6 & 76.5 & 1.1\% & 75.6 & 76.5 & 1.1\% \\ \hline
		\end{tabular}
	\end{center}
	
\end{table*}

\begin{table*}[!tb]
\caption{TA (\%) of the proposed approach against the baseline for 100 training images per class of each species.}
	\label{tab:accuracy100}
	\begin{center}
		\begin{tabular}{|c|c|c|ccc|ccc|ccc|}
			\hline
			\multirow{3}{*}{Species} & \multirow{3}{*}{CNN} & \multirow{3}{*}{Similarity} & \multicolumn{9}{c|}{Classes} \\ \cline{4-12} 
			&  &  & \multicolumn{3}{c|}{3} & \multicolumn{3}{c|}{4} & \multicolumn{3}{c|}{5} \\ \cline{4-12} 
			&  &  & Baseline & P & Gain & Baseline & P & Gain & Baseline & P & Gain \\ \hline \hline
			\multirow{2}{*}{\textit{Bird}} & ResNet18 & 81.8 & 92.0 & 93.5 & 1.5\% & 91.0 & 92.5 & 1.5\% & 90.7 & 92.1 & 1.4\% \\
			& VGG19 & 80.8 & 91.4 & 91.8 & 0.2\% & 91.0 & 91.5 & 0.5\% & 90.4 & 91.2 & 1.2\% \\
			\multirow{2}{*}{\textit{Fruit}} & ResNet18 & 72.5 & 80.6 & 80.3 & -0.3\% & 80.4 & 80.1 & -0.3\% & 80.2 & 80.0 & -0.2\% \\
			& VGG19 & 72.7 & 80.0 & 80.3 & 0.4\% & 79.3 & 80.0 & 0.9\% & 79.5 & 80.2 & 0.9\% \\
			\multirow{2}{*}{\textit{Flower}} & ResNet18 & 64.8 & 72.5 & 74.0 & 2.0\% & 72.4 & 73.3 & 1.2\% & 72.5 & 73.7 & 1.7\% \\
			& VGG19 & 63.7 & 70.7 & 71.1 & 0.5\% & 70.6 & 71.0 & 0.6\% & 70.4 & 71.8 & 2.0\% \\
			\multirow{2}{*}{\textit{Pepper}} & ResNet18 & 50.9 & 62.2 & 64.3 & 3.3\% & 62.1 & 64.2 & 3.2\% & 62.1 & 64.3 & 3.5\% \\
			& VGG19 & 52.2 & 62.6 & 62.4 & -0.3\% & 62.4 & 62.3 & -0.1\% & 62.3 & 62.5 & 0.3\% \\ \hline
			\multicolumn{2}{|c|}{Average} & 67.4 & 76.5 & 77.4 & 1.0\% & 76.2 & 76.9 & 0.9\% & 76.0 & 77.0 & \underline{\textbf{1.5}\%}
			\\ \hline
		\end{tabular}
	\end{center}
	
\end{table*}

\begin{table*}[!tb]
\caption{TA (\%) of the proposed approach against the baseline for 200 training images per class of each species.}
	\label{tab:accuracy200}
	\begin{center}
		\begin{tabular}{|c|c|c|ccc|ccc|ccc|}
			\hline
			\multirow{3}{*}{Species} & \multirow{3}{*}{CNN} & \multirow{3}{*}{Similarity} & \multicolumn{9}{c|}{Classes} \\ \cline{4-12} 
			&  &  & \multicolumn{3}{c|}{3} & \multicolumn{3}{c|}{4} & \multicolumn{3}{c|}{5} \\ \cline{4-12} 
			&  &  & Baseline & P & Gain & Baseline & P & Gain & Baseline & P & Gain \\ \hline \hline
			\multirow{2}{*}{\textit{Bird}} & ResNet18 & 81.8 & 92.4 & 93.9 & 1.5\% & 91.8 & 92.9 & 1.1\% & 91.7 & 92.5 & 0.8\% \\
			& VGG19 & 80.8 & 91.1 & 91.8 & 0.7\% & 90.9 & 91.5 & 1.4\% & 90.5 & 91.0 & 0.5\% \\
			\multirow{2}{*}{\textit{Fruit}} & ResNet18 & 72.5 & 81.0 & 80.8 & -0.2\% & 80.7 & 80.2 & -0.6\% & 80.7 & 80.2 & -0.6\% \\
			& VGG19 & 72.7 & 80.4 & 81.4 & 1.2\% & 80.3 & 81.0 & 0.9\% & 80.4 & 81.4 & 1.3\% \\
			\multirow{2}{*}{\textit{Flower}} & ResNet18 & 64.8 & 72.8 & 74.2 & 1.9\% & 72.7 & 74.2 & 2.0\% & 72.7 & 74.2 & 2.1\% \\
			& VGG19 & 63.7 & 71.9 & 72.0 & 0.2\% & 71.8 & 72.0 & 0.3\% & 71.4 & 72.0 & 0.9\% \\
			\multirow{2}{*}{\textit{Pepper}} & ResNet18 & 50.9 & 63.9 & 64.5 & 1.0\% & 63.7 & 64.2 & 0.7\% & 63.4 & 64.4 & 1.6\% \\
			& VGG19 & 52.2 & 63.7 & 63.4 & -0.4\% & 63.6 & 63.3 & -0.5\% & 63.4 & 63.2 & -0.3\% \\ \hline
			\multicolumn{2}{|c|}{Average} & 67.4 & 77.1 & 78.0 & 0.8\% & 76.9 & 77.5 & \underline{0.7\%} & 76.8 & 77.4 & 0.8\%
			\\ \hline
		\end{tabular}
	\end{center}
	
\end{table*}

\section{Conclusion}
\label{ssec:subhead}
A new transfer learning approach using the classification layer's output features (1000 dimension) is proposed in this work. We empirically examine and compare classification performance of baseline and proposed technique. Considering the training time, baseline approach lags far behind proposed approach. In addition, proposed approach outperforms the baseline technique in average. The impact of quantity and nature of training sets in classification outcomes are established by our designed RQs to prove classification layer features are transferable. We hope, our thorough investigation will help researchers to formulate best practices for efficient use of proposed strategy. In future, we would want to explore transferability of classification layer's output features in other visual tasks, for example, object detection, recognition, image captioning, etc with more classes in datasets.

{\small
\bibliographystyle{IEEEbib}
\bibliography{strings}
}
\end{document}